\pdfoutput=1

\documentclass[11pt]{article}

\usepackage{latex/acl}
\usepackage{times}
\usepackage{latexsym}

\usepackage[T1]{fontenc}

\usepackage[utf8]{inputenc}
\usepackage{kotex}

\usepackage{microtype}
\usepackage{inconsolata}
\usepackage{graphicx}
\usepackage{amsmath,amssymb,amsfonts,amsthm,mathtools}
\usepackage{algorithm}
\usepackage{algorithmicx}
\usepackage{graphicx}
\usepackage{multicol}
\usepackage{caption}
\usepackage{footnote}
\usepackage{tabularx}
\usepackage{dsfont}
\usepackage{multirow}
\usepackage{tipa}
\usepackage{comment}
\usepackage{graphicx}
\usepackage{subcaption}
\title{Interventional Speech Noise Injection \\ for ASR Generalizable Spoken Language Understanding}

\author{Yeonjoon Jung\textsuperscript{$\spadesuit\heartsuit$} \quad
        Jaeseong Lee\textsuperscript{$\spadesuit$} \quad
        Seungtaek Choi\textsuperscript{$\clubsuit$} \\
        \textbf{Dohyeon Lee\textsuperscript{$\spadesuit$} \quad
        Minsoo Kim\textsuperscript{$\spadesuit\heartsuit$} \quad
        Seung-won Hwang~\textsuperscript{$\spadesuit\heartsuit$}\thanks{Corresponding author.}} \\
        \textsuperscript{$\spadesuit$}{\small  Seoul National University}   \quad \textsuperscript{$\clubsuit$}{\small Yanolja}\\ 
        \textsuperscript{${\heartsuit}$}{\small Interdisciplinary Program in Artificial Intelligence, Seoul National University} \\
        {\small \tt \{y970120, tbvj5914, waylight, minsoo9574, seungwonh\}@snu.ac.kr} \quad
        {\small \tt seungtaek.choi@yanolja.com} 
        \\
  }

\begin{document}
\maketitle

\newtheorem{example}{Example}

\newcommand\Tstrut{\rule{0pt}{2.2ex}}       
\newcommand\Bstrut{\rule[-0.6ex]{0pt}{0pt}} 
\newcommand{\TBstrut}{\Tstrut\Bstrut} 

\newcommand{\todoc}[2]{{\textcolor{#1}{\textbf{#2}}}}
\newcommand{\jslee}[1]{{\color{purple}#1}}
\newcommand{\todored}[1]{\todoc{red}{\textbf{#1}}}
\newcommand{\todoblue}[1]{\todoc{blue}{\textbf{[[#1]]}}}
\newcommand{\todogreen}[1]{\todoc{green}{\textbf{[[#1]]}}}
\newcommand{\todoorange}[1]{\todoc{orange}{\textbf{[[#1]]}}}
\newcommand{\todopurple}[1]{\todoc{purple}{\textbf{[[#1]]}}}

\newcommand{\kjh}[1]{\todored{jihyuk: #1}}
\newcommand{\mj}[1]{\todoblue{MJ: #1}}
\newcommand{\yj}[1]{\todoblue{YJ: #1}}
\newcommand{\sw}[1]{\todoorange{SW: #1}}
\newcommand{\hist}[1]{\todoorange{hist: #1}}
\newcommand{\ms}[1]{\todopurple{MS: #1}}

\newcommand{\se}{{\it SE}}%
\newcommand{\eg}{{\it e.g.}}%
\newcommand{\ie}{{\it i.e.}}%
\newcommand{\etal}{{\it et al.}}%
\newcommand{\etc}{{\it etc}}%
\newcommand{\ours}{{WISE}}%

\newcommand{\argmin}{\operatornamewithlimits{argmin}}
\newcommand{\argmax}{\operatornamewithlimits{argmax}}
\definecolor{yellow-green}{rgb}{0.3, 0.5, 0.0}

\def\geotextual{{spatial-keyword}}
\def\geospatial{geo-spatial}
\def\PI{\mathcal{P}}
\newcommand{\XXP}[1]{{\PI(#1)}}
\def\XXQEO{\emph{$Q_1$}}
\def\kNN{\textsc{$k$NN}}
\def\XXD{\mathcal{D}}
\def\XXT{\mathcal{T}}
\newcommand{\XXDN}[0]{{D}}
\newcommand{\XXTN}[0]{{T}}
\def\Base{\textsc{Base}}
\def\TopK{\textsc{Top-$k$}}
\def\tag{{keyword}}
\def\Query{{Query}}
\newcommand{\ttag}[1]{{`#1'}}

\newcommand{\base}{\textsf{NER}}
\newcommand{\baseHash}{\textsf{NER+Hash}}

\newcommand{\mcal}[1]{{\cal{#1}}}
\newcommand{\calA}{\mbox{${\cal A}$}}
\newcommand{\calB}{\mbox{${\cal B}$}}
\newcommand{\calC}{\mbox{${\cal C}$}}
\newcommand{\calD}{\mbox{${\cal D}$}}
\newcommand{\calE}{\mbox{${\cal E}$}}
\newcommand{\calF}{\mbox{${\cal F}$}}
\newcommand{\calG}{\mbox{${\cal G}$}}
\newcommand{\calH}{\mbox{${\cal H}$}}
\newcommand{\calI}{\mbox{${\cal I}$}}
\newcommand{\calJ}{\mbox{${\cal J}$}}
\newcommand{\calK}{\mbox{${\cal K}$}}
\newcommand{\calL}{\mbox{${\cal L}$}}
\newcommand{\calM}{\mbox{${\cal M}$}}
\newcommand{\calN}{\mbox{${\cal N}$}}
\newcommand{\calO}{\mbox{${\cal O}$}}
\newcommand{\calP}{\mbox{${\cal P}$}}
\newcommand{\calQ}{\mbox{${\cal Q}$}}
\newcommand{\calR}{\mbox{${\cal R}$}}
\newcommand{\calS}{\mbox{${\cal S}$}}
\newcommand{\calT}{\mbox{${\cal T}$}}
\newcommand{\calU}{\mbox{${\cal U}$}}
\newcommand{\calV}{\mbox{${\cal V}$}}
\newcommand{\calW}{\mbox{${\cal W}$}}
\newcommand{\calX}{\mbox{${\cal X}$}}
\newcommand{\calY}{\mbox{${\cal Y}$}}
\newcommand{\calZ}{\mbox{${\cal Z}$}}
\begin{abstract}
Recently, pre-trained language models (PLMs) have been increasingly adopted in spoken language understanding (SLU). 
However, automatic speech recognition (ASR) systems frequently produce inaccurate transcriptions, leading to noisy inputs for SLU models, which can significantly degrade their performance. 
To address this, our objective is to train SLU models to withstand ASR errors by exposing them to noises commonly observed in ASR systems, referred to as ASR-plausible noises. 
Speech noise injection (SNI) methods have pursued this objective by introducing ASR-plausible noises, but we argue that these methods are inherently biased towards specific ASR systems, or ASR-specific noises.
In this work, we propose a novel and less biased augmentation method of introducing the noises that are plausible to any ASR system, by cutting off the non-causal effect of noises.
Experimental results and analyses demonstrate the effectiveness of our proposed methods in enhancing the robustness and generalizability of SLU models against unseen ASR systems by introducing more diverse and plausible ASR noises in advance. 
\end{abstract}
\section{Introduction}
\begin{figure}[!t]
	\centering
	\includegraphics[width=\linewidth]{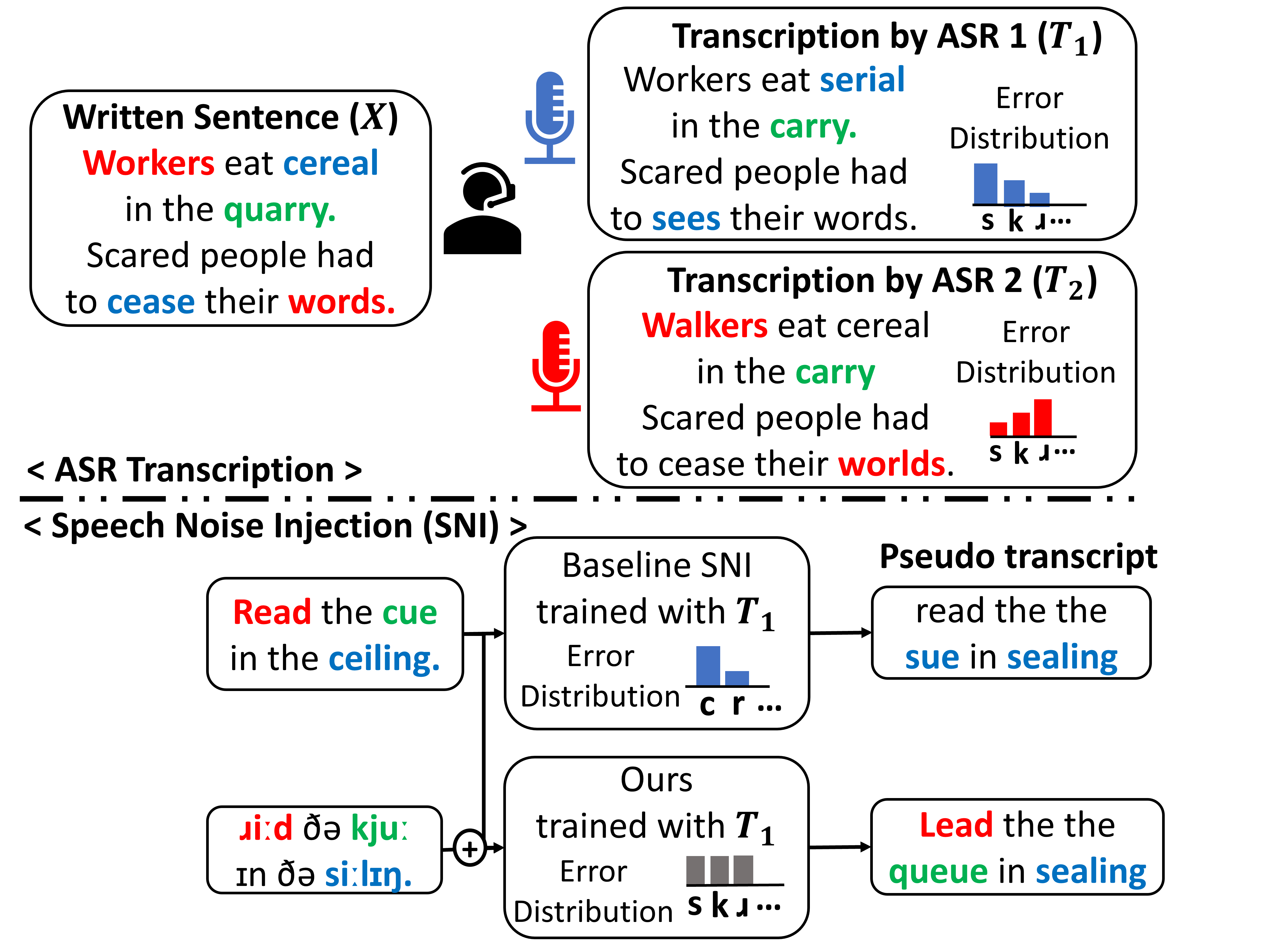}
\caption{Different ASR systems generate different ASR errors (ASR$_1$ : \textcolor{blue}{blue}, ASR$_2$ : \textcolor{red}{red}, common, or, ASR$_*$ : \textcolor{green}{green}). Biased toward a specific ASR, ASR$_1$, baseline SNI generates noises plausible only for ASR$_1$, or even some noise
that are not plausible to any (cue to sue).
Our distinction is
1) removing its bias to a specific ASR (Read to Lead), and 2) generating ASR$_*$-plausible noises (cue to queue).
}

	\label{fig:example}
 \vspace{-3mm}
\end{figure}
Pre-trained language models (PLMs) have demonstrated a robust contextual understanding of language and the ability to generalize across different domains. 
Thus, PLMs have gained widespread acceptance and been employed within the realm of spoken language understanding (SLU), such as voice assistants~\cite{broscheit-etal-2022-distributionally, 8907842}. 
A concrete instance of this application is a pipeline where an automatic speech recognition (ASR) system first transcribes audio inputs, and the transcriptions are then processed by PLMs for downstream SLU tasks~\cite{asrglue}.

However, such pipelines encounter challenges when faced with inaccuracies from ASR systems. 
We refer to these inaccuracies as ASR error words, which are phonetically similar but semantically unrelated~\cite{ruan20_interspeech,9054689}. 
For instance, as shown in Fig.~\ref{fig:example}, ASR systems might confuse words like ``cereal'' and ``serial'' or ``quarry'' and ``carry'', resulting in incorrect transcriptions such as ``workers eat serial in the carry''. 
Despite their phonetic resemblance, these ASR error words convey unintended meanings, thereby impeding the semantic understanding of speech in SLU tasks~\cite{belinkov2018synthetic}.

A well-known solution is speech noise injection (SNI) which generates likely incorrect transcriptions, namely pseudo transcriptions, then exposes PLMs to the generated pseudo transcriptions while training SLU tasks~\cite{heigold-etal-2018-robust,di-gangi-etal-2019-robust}.
Thus, the injected noises should be ASR-plausible, being likely to be generated by ASR systems from real audio input, for which the traditional methods attempted to replicate the ASR errors from written text. However, their ASR-plausibility is conditional on a particular ASR system, or ASR$_i$, where the error distribution from the collected transcriptions follows only the observed error distribution~\cite{confusion_matrix,confusion_matrix2,noisygen}.

However, different ASR systems have distinct error distributions~\cite{asr_pattern1}, which hinders the trained SLU models to be used with other ASR systems, ASR$_j$.
A straightforward solution to this issue might be to employ multiple ASR systems or to build multiple SLU models, but this approach incurs significant overheads and cannot account for the diversity of real-world ASR errors, \eg, errors due to environmental sounds. 
Instead, we investigate a novel but easier solution of introducing better generalizable noises from a single ASR system.

For this purpose, we first identify the gap between the ASR transcription in SLU and SNI
from a causality perspective.
First, SLU tasks aim to handle real audio input, where written ground-truths(GTs) are recorded as audio by humans, such that ASR errors are causally affected by recorded audio, as shown in Figure~\ref{fig:fig2_subfig1}. 
However, SNI,
when replicating error patterns, may discard the correctly transcripted texts, biased towards the observed errors.
Inspired by causality literature, we introduce two technical contributions: 1) interventional noise injection and 2) phoneme-aware generation. 
Specifically, we adopt $do$-calculus to intervene the ``revised'' in SNI to
deviate from the observed error distribution, thereby broadening the error patterns in the resulting pseudo transcripts.
For instance, as shown in Fig.~\ref{fig:example}, our SNI model can corrupt the word `Read' with the phoneme `\textturnr', which was corrupted in ASR$_2$ but not in ASR$_1$, in addition to the errors introduced by baseline SNI.

Next, we ensure that the debiased noises are plausible for any ASR system, referred to as ASR$_*$. 
This involves making GT words and ASR noises phonetically similar based on the common characteristics shared by ASR$_*$~\cite{serai2022hallucination}. 
Along with the textual input, we incorporate information on how words are pronounced. 
By being aware of pronunciation, we can introduce ASR noises that are plausible regardless of the specific ASR system used, making them ASR$_*$-plausible.

Experiments were conducted in an ASR zero-shot setting, where SNI models were trained on ASR$_i$ and tested on SLU tasks using another ASR system, ASR$_j$, on the DSTC10 Track2 and ASR GLUE benchmarks. 
Results show that our proposed methods effectively generalize across different ASR systems, with performance comparable to, or even exceeding, the in-domain setting where ASR$_j$ is used to train the SNI model.

\section{Related Work}

\subsection{SNI}
Previously proposed methods can be broadly categorized into three main approaches. 
The first, a Text-to-Speech (TTS)-ASR pipeline~\cite{tts_asr, tts_asr2}, uses a TTS engine to convert text into audio, which is then transcribed by an ASR system into pseudo transcriptions. 
However, this method struggles due to different error distributions between human and TTS-generated audio, making the pseudo transcriptions less representative of actual ASR errors. 
The second approach, textual perturbation, involves replacing words in text with noise words using a scoring function that estimates how likely an ASR system is to misrecognize the words, often employing confusion matrices~\cite{confusion_matrix3, confusion_matrix5} or phonetic similarity functions~\cite{li-specia-2019-improving,tsvetkov-etal-2014-augmenting}. 
The third method, auto-regressive generation, utilizes PLMs like GPT-2 or BART to generate text that mimics the likelihood of ASR errors in a contextually aware manner, producing more plausible ASR-like noise~\cite{noisygen}.

We consider auto-regressive noise generation as our main baseline as it has shown superior performance over other categories~\cite{asrglue,dstc10_2}
However, auto-regressive noise generation is biased to ASR$_i$, limiting the SLU model used for ASR$_i$. 
Our distinction is generalizing SNI so that the SLU tasks can be conducted with ASR$_*$.
We provide a more detailed explanation of each category in Appendix~\ref{sec:rel_work}.

\subsection{ASR Correction}
As an alternative to SNI, ASR correction
aims to denoise (possibly noisy) ASR transcription $T_i$ into $X$:
Due to its similarity to SNI, similar methods, such as
 textual perturbation~\cite{leng-etal-2021-fastcorrect-2} and auto-regressive generation~\cite{robart, chen2024hyporadise}, were used.
Also, PLMs showed impressive results~\cite{robart}, as the ASR noise words can be easily detected by PLMs due to their semantic irrelevance.
Using such characteristics, the constrained decoding method which first detects ASR errors, then corrects detected ASR errors is proposed~\cite{constdecoder}.
However, the innate robustness of PLMs~\cite{heigold-etal-2018-robust} makes SNI outperforms ASR correction~\cite{asrglue}.
In addition, it introduces additional latency in the SLU pipeline for ASR correction before the SLU model to conduct SLU tasks.
In voice assistant systems like Siri, such additional latency is crucial as the SLU pipeline should deliver responses with minimal delay to ensure real-time interaction.
Therefore, we focus on SNI for its effectiveness and minimal latency in SLU pipeline.

\section{Method}
Before introducing ISNI, we first formally define the problem of SNI and outline its causal diagram. 
Following this, we provide an overview of ISNI and detail the methods used to generate pseudo transcriptions that enhance the robustness of SLU tasks against ASR$_*$.

\subsection{Problem Formulation}
To robustify PLMs against ASR errors in SLU, the SNI model mimics the transcriptions of ASR$_i$ which can be defined as a functional mapping $\textit{F}_i: X \rightarrow T_i$.
Given the written GT $X=\{x^{k}\}_{k=1}^{n}$ with $n$ words, the SNI model $\textit{F}_i$ outputs the pseudo transcription $T_i=\{t^{k}_{i}\}_{k=1}^{n}$, simulating the transcriptions produced by ASR$_i$. 
These pseudo transcriptions, $T_i$, are subsequently used to train SLU models. 
For each word $x^{k}$, $z^{k} \in Z$ indicates whether ASR$_i$ makes errors ($z^{k}=1$, hence $x^{k} \neq t^{k}_{i}$) or not ($z^{k}=0$, hence $x^{k}=t^{k}_{i}$). 
However, noises generated by this model differ from those of the other ASR system, ASR$_j$, and SLU model trained with the generated noises struggles with the errors from ASR$_j$.
Therefore, we propose building an SNI model, $\textit{F}_*$, capable of generating ``ASR$_*$-plausible'' pseudo transcripts $T_*$, which are plausible for any ASR system, ASR$_*$.

\subsection{Causality in SNI}
\begin{figure}[tbp]
    \centering
    \begin{subfigure}[t]{0.15\textwidth}
        \centering
        \includegraphics[width=\textwidth]{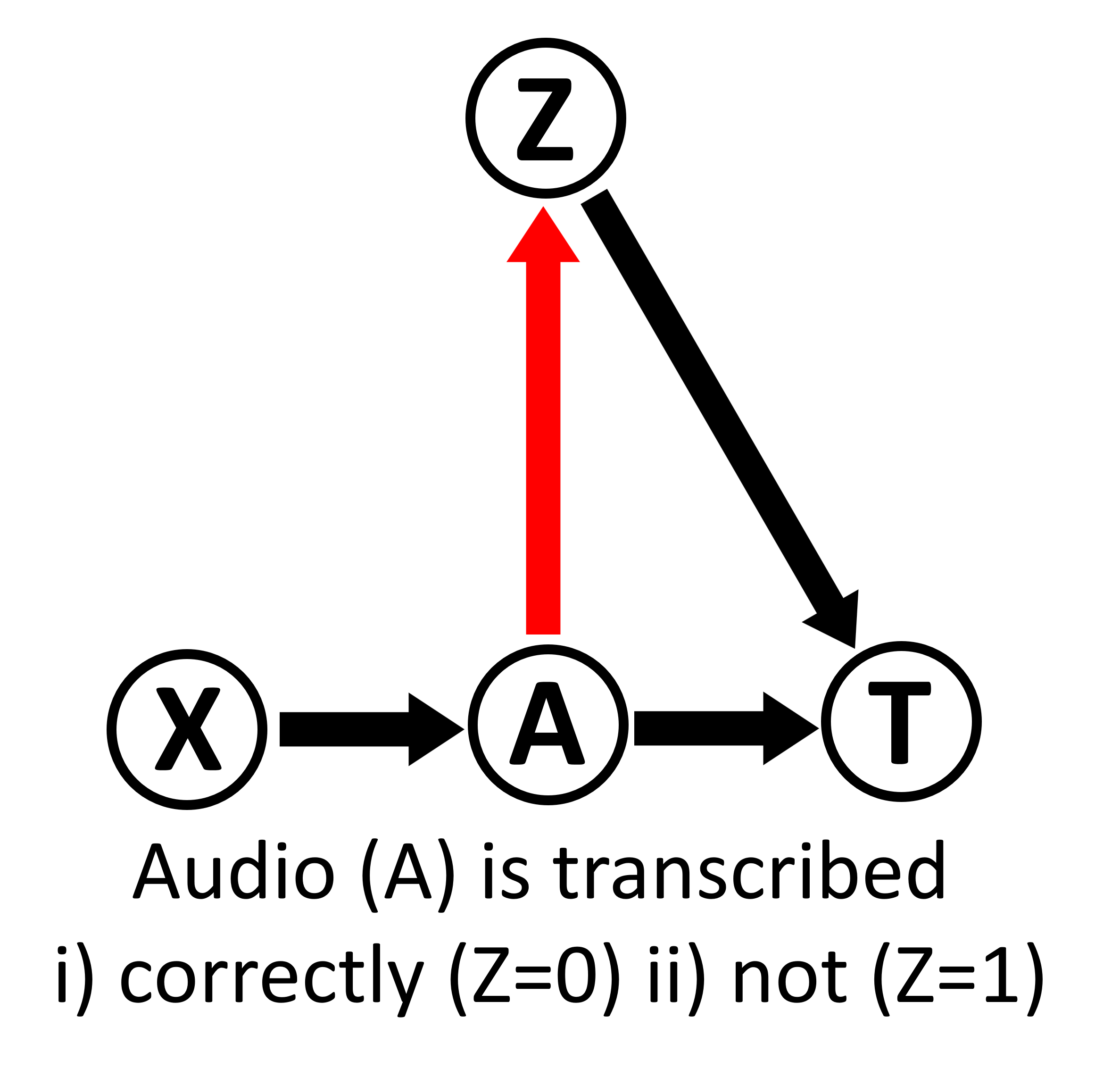}
        \caption{Causal graph of ASR transcription. $X$ causally influences to $Z$ through directed path.}
        \label{fig:fig2_subfig1}
    \end{subfigure}
    \hfill
    \begin{subfigure}[t]{0.135\textwidth}
        \centering
        \includegraphics[width=\textwidth]{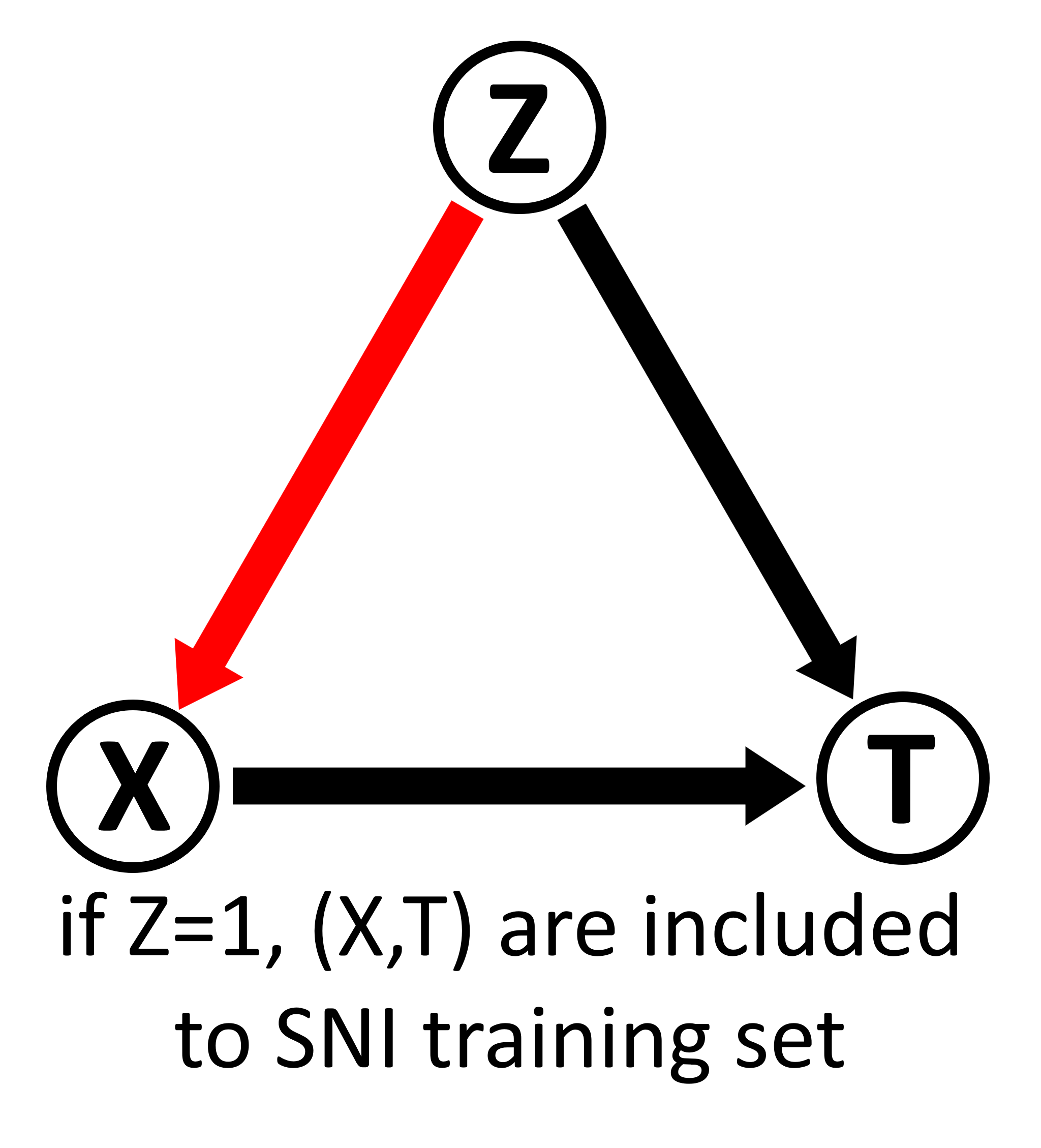}
        \caption{Causal graph of SNI training data collection. $X$ and $Z$ are non-causally related through backdoor path.}
        \label{fig:fig2_subfig2}
    \end{subfigure}
    \hfill
    \begin{subfigure}[t]{0.135\textwidth}
        \centering
        \includegraphics[width=\textwidth]{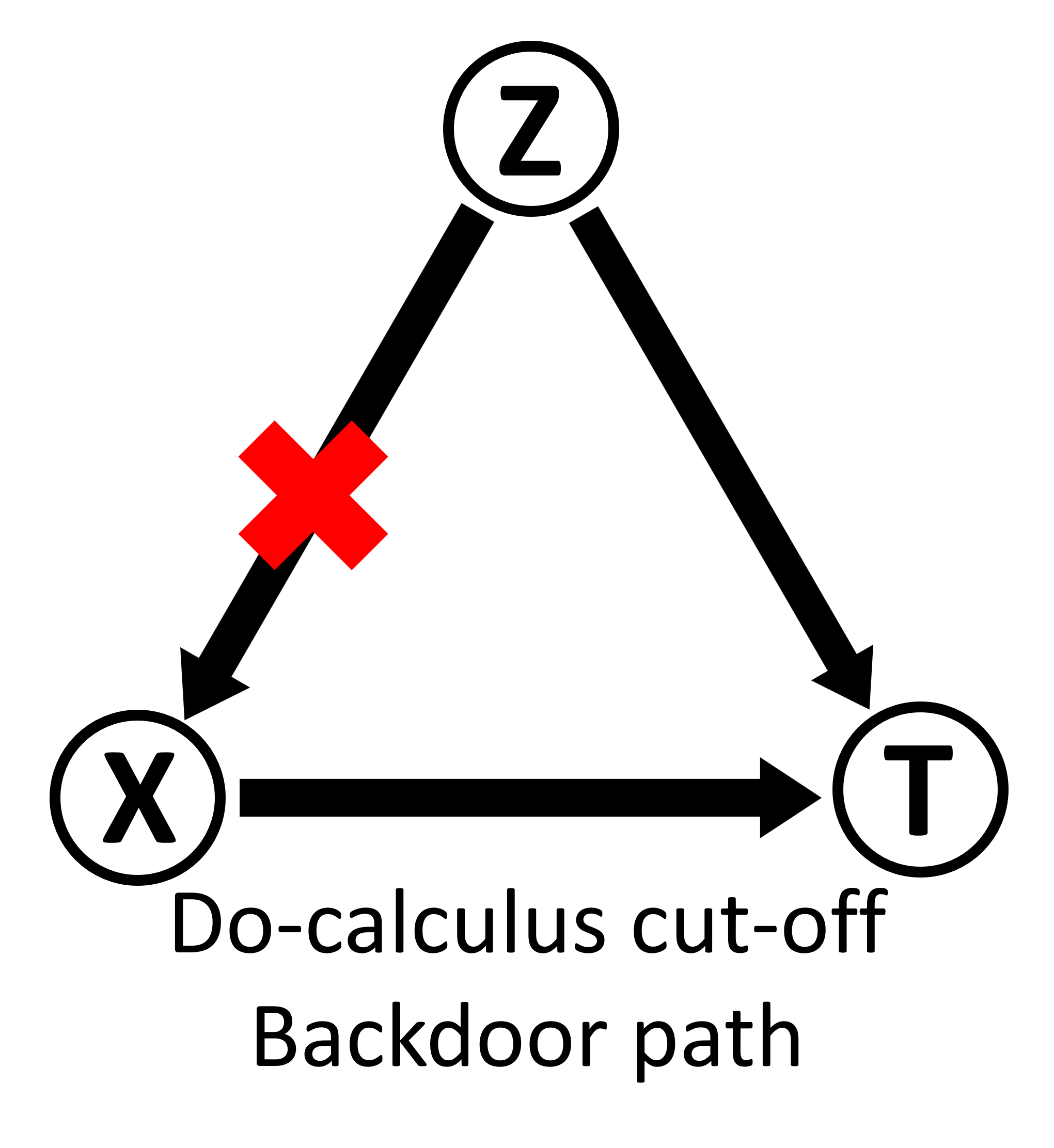}
        \caption{Causal graph of ISNI adopting $do$-calculus. Path between $Z$ and $X$ are cut-off} 
        \label{fig:fig2_subfig3}
    \end{subfigure}
\caption{The causal graph between ASR transcription (\subref{fig:fig2_subfig1}), SNI training data generation (\subref{fig:fig2_subfig2}), and ISNI (\subref{fig:fig2_subfig3})}
  \label{fig:fig2}
\end{figure}

To achieve this, we compare the underlying causal relations between the transcription process in SLU and SNI training data generation, which are depicted in Fig.~\ref{fig:fig2_subfig1}. 
During ASR transcription, written GT $X$ is first spoken as audio $A$, which is then transcribed as transcription $T$ by ASR$_i$.
Depending on the audio $A$, ASR$_i$ may make errors ($Z=1$) or not ($Z=0$) in the transcription $T$. 
While transcribing, $X$ influences $Z$ through causal paths where every edge in the path is directed toward $Z$, where $Z$ acts as a mediator between $X$ and $Z$.

In SNI, the $X$ and $T$ transcribed by ASR$_i$ are filtered based on $Z$ since they do not exhibit the error patterns required for training.
However, such filtering induces a backdoor path and a non-causal relation in SNI. 
To further elucidate the causal relations in SNI training data generation depicted in Fig.~\ref{fig:fig2_subfig2}, we outline the causal influences as follows:
\begin{itemize}
\item $X \rightarrow T$: There is a direct causal relationship where the clean text $X$ influences the transcribed text $T$.
\item $Z \rightarrow T$: If $z^{k} \in Z$ is 1 (an error occurs), it directly affects the corresponding transcription $t^{k}_{i}$, causing it to deviate from the clean text $x^{k}$.
\item $Z \rightarrow X$: In the SNI training data collection process, $Z$ determines if $X$ is included. This means that only when the ASR system makes a mistake, indicated by any value $z^k \in Z$ being 1, the corresponding text is included in the training data. So, errors by the ASR system decide which clean texts are chosen.
\end{itemize}
The backdoor path $X \leftarrow Z \rightarrow T$, while ensuring that only instances where ASR$_i$ made errors are included, introduces bias in previous SNI models based on conventional likelihood, defined as: 
\begin{equation}
P(t^{k}_{i}|x^{k})=\sum_{z^{k}}P(t^{k}_{i}|x^{k},z^{k})P(z^{k}|x^{k}).
\label{eq:likelihood}
\end{equation}
In contrast to ASR transcription where $z^{k}$ is a consequence of $x^{k}$ and the mediator between $x^{k}$ and $t^{k}$, $z^{k}$ conversely influences $x^{k}$  and acts as a confounder.
Thus, the influence of $x^k$ to $t^k$ is drawn from $z^k$, not from $x^k$, thereby distorting the causal effect from $X$ to $T$. 
Such distortion skews the prior probability $P(z^{k}|x^{k})$ so that texts frequently incorrectly transcribed by ASR$_i$ are also noised by SNI.

\subsection{$do$-calculus for ISNI}
\label{sec:do_calculus}

To mitigate biases in SNI toward the ASR$_i$, we propose interventional SNI (ISNI) where the non-causal relations are cut off by adopting $do$-calculus.
Adopting $do$-calculus for conventional likelihood in Eq.~\ref{eq:likelihood}, ISNI can be formulated as follows:
\begin{equation}
    P(t^{k}|do(x^{k}))= \sum_{z^{k}}P(t^{k}|x^{k},z^{k})\cdot P(z^{k}).
    \label{eq:do}
\end{equation}  
Compared to Eq.\ref{eq:likelihood},  the prior probability $P(z^{k}|x^{k})$ is replaced with $P(z^{k})$.
This difference implies that the non-causal path $Z \rightarrow X$ is cut off, as the influence of $x^k$ on $t^k$ is not drawn from $z^k$ in Eq.~\ref{eq:do} as prior probability $P(z^k)$ is estimated independently from the non-causal path $Z \rightarrow X$. 
Thus, the influence of $x^k$ to $t^k$ is induced solely from $x^k$.
We provide more detailed proof of Eq.~\ref{eq:do} in Appendix~\ref{sec:proof_do}.

\subsection{Overview of Our Proposed Approach}
\label{sec:overview}
\begin{figure}[t]
	\centering
	\includegraphics[width=1.0\linewidth]{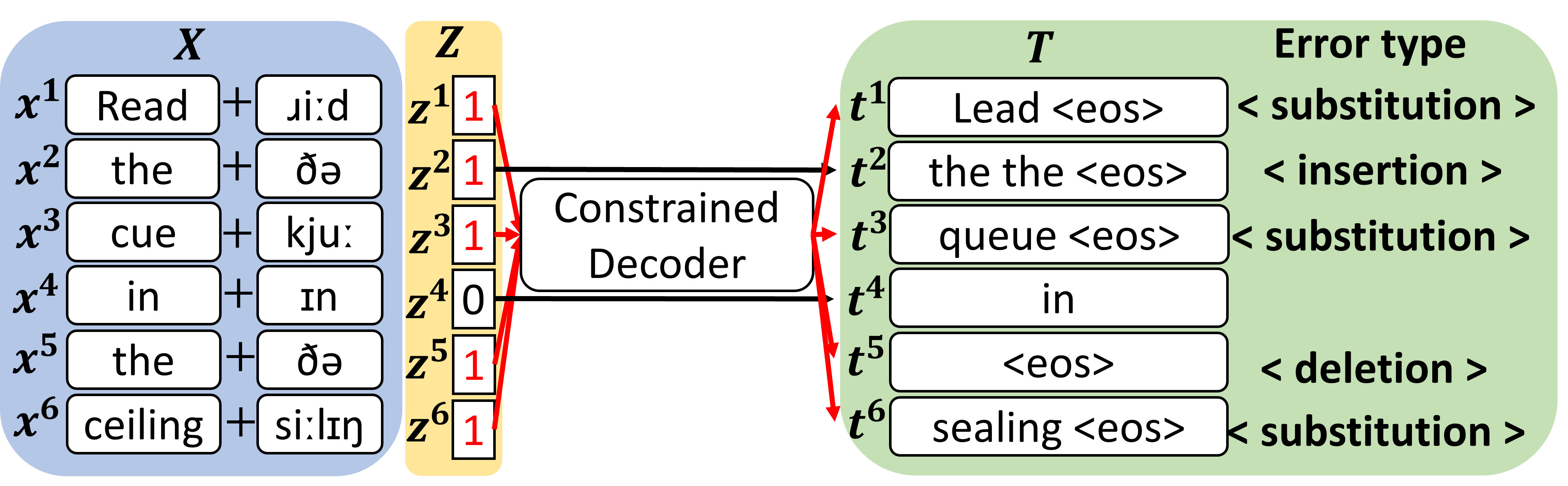}
\caption{Overview of ISNI. ISNI generates ASR noise word $t^k$ for the clean text $x^k$ whose corresponding $z^k$ is 1. The error type of $t^k$ is determined by the generated output.}
	\label{fig:fig3}
\end{figure}

In this section, we briefly overview how ISNI is implemented to generate ASR$_*$-plausible noises of different error types, as illustrated in Fig.~\ref{fig:fig3}.
To achieve this, ISNI implements two distinct terms in Eq.~\ref{eq:do}: $P(z^{k})$ and $P(t^{k}|x^{k}, z^{k})$.

For debiased noise generation $P(z^k)$, it is essential to cut off the influence of $x^k$ on $z^k$ so that $z^k$ becomes independent of $x^k$. 
ISNI explicitly controls $z^k$ and removes the dependence on $x^k$ by implementing SNI with constrained decoding, following \citet{constdecoder}.
Specifically, by utilizing \textit{do}-calculus? to determine $z$, ISNI introduces corruption in tokens that  ASR$_i$ transcribes correctly but ASR$_j$ may mistranscribe, thereby adding these as noise in the pseudo transcripts.
For example, the token $x$ with $z=1$, such as `cue' in Fig.~\ref{fig:fig3}, is fed to the constrained decoder.
The constrained decoder then generates the noise word $t$ to replace $x$.

The noise word $t$ can be categorized as three conventional ASR error types: substitution, insertion, and deletion~\cite{jurafsky2019speech}, which is determined by the generation output of ISNI.
Substitution errors occur when the generated output is a new token that replaces $x$. 
Insertion errors occur when the constrained decoder outputs multiple tokens for $x$. 
For example, in Fig.~\ref{fig:fig3}, the constrained decoder outputs two `the' words for $x^2$, which means the insertion of the newly generated `the' beside the given word `the'. 
Deletion errors occur when the constrained decoder outputs only the $eos$ token, representing the end of generation. 
If the generation ends without any tokens, as in $t^5$, the empty string replaces $x^5$. 
ISNI generates these error types synthetically, as demonstrated in the quantitative study in Appendix~\ref{sec:quantitiate_study} and the generated examples in Appendix~\ref{sec:pseudo_transcripts_examples}.

For ASR$_*$-plausible generation $P(t^{k}|x^{k},z)$, we provide the phonetic information of $x$, such as `kju' in Fig.~\ref{fig:fig3}, to ISNI to understand how $x$ is pronounced. 
The constrained decoder evaluates the generation probability based on phonetic similarity, recognizing that `c' is pronounced as the phoneme `k'. 
Consequently, ISNI generates the ASR$*$-plausible noise word `queue' which is phonetically similar to `cue'.

\subsection{Interventional SNI for Debiasing}
\label{sec:interventional_sni}
To mitigate bias from dependence on ASR$_i$, we propose an intervention on the non-causal effect $X \leftarrow Z$.   
This approach diversifies the noise words in the pseudo transcripts, including those rarely corrupted by ASR$_i$. 
By applying \textit{do}-calculus as shown in Eq.~\ref{eq:do}, we derive a following equation for SNI:
\begin{equation}
    P(t^{k}|do(x^{k})) = \sum_{z}P(t^{k}|x^{k},z) \cdot P(z).
                                \label{eq:do-calculus}
\end{equation}
This intervention ensures that the corruption of words $x^k$ in the pseudo transcripts does not depend on any particular ASR system.
Consequently, it allows for the corruption of words that are typically less susceptible to errors under ASR$_i$.

For each word $x^k$,  the prior probability $P(z)$ in Eq.~\ref{eq:do-calculus} represents the probability that any given word $x^k$ will be corrupted.
To simulate $P(z)$, we assume that the average error rates of any ASR system, ASR$_*$ would be equal for all words and sample a random variable $a^k$ for each word $x^k$ from a continuous uniform distribution over the interval $[0,1]$.
If $a^k \leq P(z)$, we set $z$ to 1, indicating an incorrect transcription of $x^k$ and generating its noise word $t^{k}$. 
Otherwise, $z$ is set to 0, and $t^{k}$ remains identical to $x^{k}$, mimicking an accurate transcription. 
We set the prior probability $P(z)$ as a constant hyperparameter for pseudo transcript generation.

To generate a noise word $t^k$ whose corruption variable $z^k$ is determined independently of any biases, we adopt a constrained generation technique that outputs $t^{k}$ to input $x^{k}$~\cite{constdecoder}.
To implement constrained generation, we use BERT~\cite{devlin-etal-2019-bert} for encoding the written GT $X$ into the vector representation $E$ as follows:
\begin{equation}
    E_{encoder} = \textit{BERT}(M_{word}(X)+M_{pos}(X)),
    \label{eq:encoder}
\end{equation}
where $e_{encoder}^{k} \in E_{encoder}$ is the encoder representation of the token $x^{k}\in X$ and $M_{word}$ and $M_{pos}$ denote word and position embeddings.

The transformer decoder~\cite{NIPS2017_3f5ee243} generates $t^{k}$ constrained on $x^k$.
To encompass all ASR error types, the transformer decoder generates multiple tokens (see Section~\ref{sec:overview}), denoted as $m$ tokens, $\{\Tilde{t}^{k}_{l}\}_{l=1}^{m}$.  
These tokens replace $x^k$ in the pseudo transcripts. 
 The error type of $t^{k}$ is contingent on $m$:
 \begin{itemize}
     \item $m=1$ corresponds to deletion, generating only $eos$ is generated; $x^k$ is substituted by the empty string. 
    \item If $m=2$, one token in addition to $eos$ is generated; this token substitutes $x^k$.
    \item When $m>2$, additional token will be inserted to $x^k$, to simulate insertion errors.
 \end{itemize}

To generate $\Tilde{t}^{k}_{l}$, the input of the decoder comprises the encoder representation of the written word $e_{encoder}^{k}$ and the special embedding vector $bos$ (beginning of sequence) and the tokens generated so far, $\{\Tilde{t}^{k}_{0},..., \Tilde{t}^{k}_{l-1}\}$ as follows:
\begin{equation}
    E_{decoder}^{k}= H_{decoder} \cdot [e_{encoder}^{k}; bos,\Tilde{t}^{k}_{0},..., \Tilde{t}^{k}_{l-1}],
    \label{eq:decoder_input}
\end{equation}
where $H_{decoder}$ is the weight matrix of the hidden layer in the decoder.
The transformer decoder then computes the hidden representation $d^{k}$ by processing the input through its layers as follows:
\begin{align}
    d^{k}= TransformerDecoder(Q,K,V),\\
    Q=E_{decoder}^{k},K,V=E_{encoder},
    \label{eq:decoder}
\end{align}
where $E_{decoder}^{k}$ serves as query $Q$, and $E_{encoder}$ is used for both key $K$ and value $V$ in the transformer decoder.
Finally, the probability of generating each token $\Tilde{t}^{k}_{l}$ is calculated using a softmax function applied to the product of the word embedding matrix $M_{word}$ and the hidden representation $d^{k}$ added to the trained bias $b_{n}$:
\begin{equation}
    P_{n}(\Tilde{t}^{k}_{l}|x^k,\Tilde{t}^{k}_{0},..., \Tilde{t}^{k}_{l-1}) = \small softmax(M_{word} \cdot d^{k}+b_{n}),
    \label{eq:gen_prob}
\end{equation}
where $b_{n}$ are trained parameters. 
Here, we note that ISNI generates different error types depending on the output as we explained in Sec.~\ref{sec:overview}.

We used the ASR transcription $T$ to train ISNI model.
ISNI is supervised to maximize the log-likelihood of the $l$-th token $\hat{t}^{k}_{l}$ from the ASR transcription as follows:
\begin{equation}
    \mathcal{L}_{n} =-\sum log(\textit{P}_{n}(\hat{t}^{k}_{l}|x^{k},\hat{t}^{k}_{0},..., \hat{t}^{k}_{l-1})).
\end{equation}

\subsection{Phoneme-aware Generation for Generalizability}
\label{phoneme_head}

The next step is to ensure the noise word generation $P(t^{k}|x^{k},z)$ ASR$_*$-plausible.
We adjust the generation probability $P_{n}$ with the phoneme-based generation probability $P_{ph}$ so that the ASR$_*$-plausible noise words can be generated as followed:
\begin{multline}
    \textit{P}_{gen}(\Tilde{t}^{k}_{l}|x^k,\Tilde{t}^{k}_{0},..., \Tilde{t}^{k}_{l-1}) = \textit{P}_{n}(\Tilde{t}^{k}_{l}|x^k,\Tilde{t}^{k}_{0},..., \Tilde{t}^{k}_{l-1}) \\ \cdot \textit{P}_{ph}(\Tilde{t}^{k}_{l}|x^k,\Tilde{t}^{k}_{0},..., \Tilde{t}^{k}_{l-1}).  
    \label{eq:phoneme_correction_prob}
\end{multline}

Our first proposal for $\textit{P}_{ph}$ is to provide the phonetic characteristics of each token via phoneme embedding $M_{ph}$.
$M_{ph}$ assigns identical embeddings to tokens sharing phonetic codes, thereby delivering how each word is pronounced.
Phoneme embedding $M_{ph}$ is incorporated into the input alongside word and position embeddings as follows:
\begin{multline}
     E_{encoder} = BERT(\lambda_{w}\cdot M_{word}(X)\\+(1-\lambda_{w})\cdot M_{ph}(X)+M_{pos}(X)),
    \label{eq:phoneme_encoder}
\end{multline}
where $\lambda_{w}$ is a hyperparameter that balances the influence of word embeddings and phoneme embeddings.
The input for the decoder is formulated similarly to Eq.~\ref{eq:decoder_input}.

Fed both the word and phoneme embedding, the decoder then can understand the phonetic information of both the encoder and decoder input.
Aggregating such information, the decoder would yield the hidden representation $d^{k}$ as in Eq.~\ref{eq:decoder}.
Then, we feed the hidden representation $d^k$ to classification head to evaluate the phoneme based generation probability $\textit{P}_{ph}$ as follows:
\begin{equation}
    \textit{P}_{ph}(\Tilde{t}^{k}_{l}|x^k,\Tilde{t}^{k}_{0},..., \Tilde{t}^{k}_{l-1}) = softmax(M_{ph} \cdot d^{k}+b_{ph}),
    \label{eq:phoneme_gen_prob}
\end{equation}
where classification head has the phoneme embedding matrix same as in Eq.~\ref{eq:phoneme_encoder} and bias $b_{ph}$.
Using the phoneme embedding $M_{ph}$ instead of the word embedding $M_{word}$ in Eq.~\ref{eq:gen_prob}, $\textit{P}_{ph}(\hat{t}^{k}|x^{k})$ can be evaluated based on the phonetic information.

Our second proposal for $\textit{P}_{ph}$ is phonetic similarity loss  $\mathcal{L}_{ph}$ supervising the phonetic information using phonetic similarity.
This approach aims to generate ASR$_*$-plausible noise words by assessing the phonetic resemblance of each token to $x^k$.
$\textit{P}_{ph}$ should evaluate how phonetically similar each token is to the $x^{k}$.
Phonetic similarity is advantageous as it quantifies the differences in phonetic codes, thereby allowing for objective supervision of $\textit{P}_{ph}(\hat{t}^{k}|x^{k})$.
We utilize the phoneme edit distance $D$, as outlined in \citet{9662078}, to calculate phonetic similarity. 
The phoneme edit distance $\textit{D}(w^{p}, w^{q})$ measures the minimum edit distance between the phonetic codes of two words, reflecting how closely one word is pronounced to another. 
Notably, $\textit{D}(w^{p}, w^{q})$ leverages articulatory features to compute similarity, which incrementally increases as the phonetic resemblance between the word pair enhances.

Phonetic similarity, $S$, is defined as follows:
\begin{equation}
    \textit{S}(w^{p}, w^{q}) = max(|C_p|-\textit{D}(w^{p}, w^{q}), 0),
\end{equation}
where $|C_p|$ is the length of the phonetic code of the word $w^p$.
This formulation ensures that $\textit{S}(w^{p}, w^{q})$ attains higher values when $w^p$ and $w^q$ are phonetically similar, and approaches zero when there is no phonetic resemblance.

To supervise $\textit{P}_{ph}$, phonetic similarity should be formulated as a probability distribution. 
For such purpose, we normalize phonetic similarity and compute the supervision $\textit{R}(\hat{t}^{k})$ as follows:
\begin{equation}
    \textit{R}(\hat{t}^{k}) = \frac{\textit{S}(t^{k},w)}{\sum_{w^{\prime}\in W}\textit{S}(t^{k},w^{\prime} )}.
 \end{equation}
Then, $\textit{P}_{ph}$ is supervised by loss defined as follows:
 \begin{equation}
     \mathcal{L}_{ph} = KL(\textit{P}_{ph}(\Tilde{t}^{k}_{l}|x^k,\Tilde{t}^{k}_{0},..., \Tilde{t}^{k}_{l-1})| \textit{R}(\hat{t}^{k})),
 \end{equation}
 where $KL$ is the KL divergence loss.
Finally, ISNI is optimized to jointly minimize the total loss $\mathcal{L}_{tot}$ which is defined as follows:
\begin{equation}
    \mathcal{L}_{tot} = \mathcal{L}_{n} + \lambda_{ph} \cdot \mathcal{L}_{ph},
    \label{eq:total_loss}
\end{equation}
where $\lambda_{ph}$ is the weight of $\mathcal{L}_{ph}$.
Supervised to evaluate the generation probability based on the phoneme information, the phoneme generation probability $\textit{P}_{ph}$ ensures the ASR$_*$-plausible noise words and $M_{ph}$ contains phonetic information.

\section{Experiments}
In this section, we delineate our experimental setup, datasets, and baseline, for SNI and SLU.
Our experiments, conducted in an ASR zero-shot setting, train SNI models on one ASR system (ASR$_i$) and test SLU models on another (ASR$_j$), to evaluate generalizability across different ASR systems.

\subsection{Dataset}
\label{sec:dataset}
\paragraph{SNI Training} For SNI training, we selected datasets consistent with our primary baseline, Noisy-Gen~\cite{noisygen}, utilizing three popular corpora: Common Voice, Tatoeba audio, and LJSpeech-1.1, with the MSLT corpus serving as the validation set. Audio recordings from these sources were processed using DeepSpeech to collect transcriptions that exhibit ASR errors. This approach yielded approximately 930,000 pairs of ground-truth and ASR-noised transcriptions.

\paragraph{SLU Training} To demonstrate ASR generalizability across diverse SLU tasks, we utilize two benchmarks: ASR GLUE and DSTC10 Track 2. 
ASR GLUE, an adaptation of the widely recognized NLU benchmark GLUE, includes two natural language inference tasks, QNLI and RTE, and one sentiment classification task, SST2. 
To simulate real-world background noises, ASR GLUE benchmark randomly injected background noise audios into speech.  
Noises are injected in 4 levels, high, medium, low, and clean.
Randomly injected noises introduce ASR errors across a broader range of phonemes.
DSTC10 Track 2, tailored for spoken language dialogue systems, comprises three subtasks: Knowledge-seeking Turn Detection (KTD), Knowledge Selection (KS), and Knowledge-grounded Response Generation (RG). 
Details of the datasets for SNI training and SLU testing are available in Appendix~\ref{sec:dataset_details}.

\subsection{Baselines}
In our study, we utilized diverse SNI models, including the GPT2-based auto-regressive SNI model, NoisyGen, as our primary baseline, given its proven efficacy in various spoken language tasks~\cite{noisygen,asrglue}. 
For the ASR GLUE benchmark, we also included NoisyGen (In Domain)~\cite{asrglue}, another GPT2-based SNI model that, unlike our standard approach, uses the same ASR system for both training and testing, thereby not adhering to the ASR zero-shot setting.
Also, PLMs trained only with written GT are used to show the necessity of exposure to ASR noises.
Demonstrating that SNI models can match or surpass NoisyGen (In Domain) will confirm their ASR generalizability.

Additionally, for the DSTC10 Track 2, we incorporated the state-of-the-art TOD-DA~\cite{tian2021tod-da} as a baseline, selected for its inclusion of both TTS-ASR pipeline and textual perturbation techniques, which are absent in NoisyGen.
We selected TOD-DA because it covers two distinct categories of SNI which were not covered by NoisyGen: TTS-ASR pipeline and textual perturbation. 

\subsection{ASR system for SNI training and SLU testing}
We provide the ASR systems used for SNI training and SLU testing in Table~\ref{tab:ASRs}.
\paragraph{SNI Training}
We chose the open-source Mozilla DeepSpeech ASR system~\cite{deepspeech} primarily because it aligns with the use of commercial ASR systems in previous SLU studies, including our main baseline, NoisyGen\footnote{For a detailed comparison of commercial ASR systems, see Appendix~\ref{sec:asr_details}.}. 
We specifically selected DeepSpeech because, as an open-source system, it provides transparency and flexibility while demonstrating a word error rate comparable to other closed commercial ASR systems.
Moreover, our decision to use DeepSpeech reflects a practical scenario where SNI models trained on one ASR system need to demonstrate robustness and adaptability when applied to newer or different ASR systems, such as LF-MMI TDNN for NoisyGen (In Domain)~\cite{asrglue} and Wave2Vec for TOD-DA~\cite{yuan2017wave2vec}.

\begin{table}[h]
\scalebox{0.7}{
\begin{tabular}{l|c|c|c|c}
\hline
ASR                  & DeepSpeech        & LF-MMI TDNN               & Wave2vec          & Unknown \\ \hline
\multirow{2}{*}{SNI} & ISNI,             & NoisyGen                  & TOD-DA            & -       \\
                     & NoisyGen          & (In Domain)               &                   &         \\ \hline
\multirow{2}{*}{SLU} & \multirow{2}{*}{-} & \multirow{2}{*}{ASR GLUE} & \multirow{2}{*}{-} & DSTC10  \\
                     &                   &                           &                   & Track2  \\ \hline
\end{tabular}
}
\caption{ASR systems for SNI training and SLU testing.}
\label{tab:ASRs}
\end{table}
\paragraph{SLU Testing} To evaluate the generalizability of SNI across various ASR systems, we employed distinct ASR systems for SLU testing in the ASR GLUE and DSTC10 Track 2 benchmarks.
As detailed in Table~\ref{tab:ASRs}, the ASR GLUE test set was transcribed using an LF-MMI TDNN-based ASR system, while an unknown ASR system was used for the DSTC10 Track 2 validation and test sets.

\begin{table*}
\centering
    \scalebox{0.7}{
\begin{tabular}{l|cccc|cccc|cccc}
\hline
Task        & \multicolumn{4}{c|}{QNLI}                                         & \multicolumn{4}{c|}{RTE}    & \multicolumn{4}{c}{SST2}    \\ \hline
Noise Level & High           & Medium         & Low            & Clean          & High & Medium & Low & Clean & High & Medium & Low & Clean \\ \hline
Written GT    & 70.22 & 73.00  & 81.45         & 90.00             & 45.84 & 48.82  & 60.71    &   78.57    & 79.11 & 80.19  & 81.41  & 93.51 \\ \hline
NoisyGen    & 71.00             & 73.34          & 79.73          &  \textbf{86.00}             & 46.43     & 50.00 & 58.33    &   \textbf{60.71}    & 80.85 &87.34 & 81.84 & \textbf{92.86} \\

ISNI (Ours)        & \textbf{76.89} & \textbf{77.89} &  \textbf{82.56}          &  \textbf{86.00}             &     \textbf{56.55} &  \textbf{58.93}      & \textbf{60.12}    &   \textbf{60.71}    &  \textbf{82.36}&  \textbf{88.12} & \textbf{85.17} &  91.56 \\ \hline
Noisy-Gen (In Domain)    & 74.00          & 77.39       & 83.44 & 88.67 &  53.57   &   58.93        & 59.52    &   60.71    & 83.98&88.75 & 84.20 & 92.86 \\\hline
\end{tabular}
}
\caption{Accuracy on QNLI and RTE, SST2 of ASR GLUE benchmark.}
\label{tab:asrglue_result}
\end{table*}

\subsection{Experimental Settings}
\paragraph{SNI Training}
For ISNI implementation, we utilized BERT-base~\cite{devlin-etal-2019-bert} as an encoder and a single Transformer decoder layer~\cite{NIPS2017_3f5ee243}, aligning with established methodologies~\cite{constdecoder}.
The balance between word and phoneme embeddings was set with $\lambda_{w}$ at 0.5 in Eq.~\ref{eq:phoneme_encoder}, and the phoneme generation loss weight $\lambda_{ph}$ was also adjusted to 0.5 (Eq.~\ref{eq:total_loss}).
We provide further details in Appendix~\ref{sec:ISNI_model_detail}.

\paragraph{SLU Training}
Utilizing the trained ISNI, we convert the written GTs into pseudo transcripts. 
During the generation of pseudo transcripts, we set the prior probability $P(z)$ for ASR GLUE at 0.15 and for DSTC10 Track2, at 0.21, based on validation set result of downstream SLU tasks\footnote{Directly matching $P(z)$ to the word error rate (WER) of the ASR system is not feasible in an ASR zero-shot setting where WER for unknown ASR systems is not available. Additionally, training with arbitrary word error rates might bias the SLU model towards certain ASR systems.}. 
To ensure a fair comparison, we set phonetic similarity thresholds in our baseline, NoisyGen, for filtering out dissimilar pseudo transcripts, based on the validation set result of downstream SLU tasks. 
In terms of downstream task models, we implemented BERT-base for ASR GLUE and GPT2-small for DSTC10 Track 2, consistent with baseline configurations~\cite{dstc10_2,asrglue}.

\section{Results}

\begin{table*}[t]
\centering
    \scalebox{0.68}{
\begin{tabular}{l|ccc|ccc|cccccccc}
\hline
Task                & \multicolumn{3}{c|}{KTD}                         & \multicolumn{3}{c|}{KS}                                                                                  & \multicolumn{8}{c}{RG}                                                                                                                                                                                                                            \\ \hline
Metric              & P              & R              & F1             & MRR@5 & R@1 & R@5 &  B@1 & B@2 & B@3 & B@4 & M              & RG@1 &  RG@2 & RGL             \\ \hline
TOD-DA              & 88.58          & 89.75          & 89.16          & 67.29                              & 60.51                            & 76.75                            & 8.31                             & 4.74                             & 2.33                             & 0.92                             & 13.02          & 16.45                            & 6.49                             & 15.13          \\
NoisyGen            & \textbf{89.42} & 90.34          & 89.88          & 65.80                               & 57.10                             & 76.62                            & 9.88                             & 5.42                             & 2.47                             & 0.97                             & 13.62          & 17.19                            & 6.42                             & 15.77          \\
ISNI (Ours)                & 88.19          & \textbf{92.97} & \textbf{90.52} & \textbf{72.61}                     & \textbf{66.43}                   & \textbf{81.11}                   & \textbf{15.32}                   & \textbf{9.79}                    & \textbf{5.10}                     & \textbf{2.36}                    & \textbf{20.18} & \textbf{24.42}                   & \textbf{10.65}                   & \textbf{22.64} \\ \hline
- phoneme-aware generation & 88.29          & 90.48          & 89.37          & 71.12                              & 64.49                            & 79.68                            & 14.61                            & 9.22                             & 4.59                             & 2.36                             & 18.84          & 22.55                            & 9.66                             & 21.33          \\
\quad -intervention       & 88.62          & 91.22          & 89.90           & 69.82                              & 63.2                             & 78.79                            & 13.77                            & 8.66                             & 4.17                             & 1.99                             & 18.66          & 22.23                            & 9.36                             & 20.89          \\ \hline
\end{tabular}
}
\caption{Results of DSTC10 Track2. Precision (P) and Recall (R), F1 is used to evaluate KTD. Mean Reciprocal Rank at 5 (MRR@5) and Recall at 1, 5 (R@1,5) are used to evaluate KS. To evaluate RG, Bleu at 1,2,3,4 (B@1,2,3,4) and Meteor (M) and Rouge at 1,2,L (RG@1,2,L) are used.}
\label{tab:dstc_results}
\end{table*}
We now present our experimental results, addressing the following research questions:

\noindent \textbf{RQ1:} Is the ASR zero-shot setting valid and how effective are ISNI in the ASR zero-shot setting?


\noindent \textbf{RQ2:} Can ISNI robustify the various SLU tasks in the ASR zero-shot setting?

\noindent \textbf{RQ3:} Does each of methods contribute to robustification?

\subsection{RQ1: Validity of ASR Zero-shot Setting and Effectiveness of ISNI. }
To demonstrate the ASR generalizability of SNI models, we compare them with NoisyGen (In Domain) on ASR GLUE in Table~\ref{tab:asrglue_result}.
Unlike Noisy-Gen and our models tested in an ASR zero-shot setting, NoisyGen (In Domain) was trained and tested using the identical ASR system, which is incompatible with the ASR zero-shot setting.

Our findings indicate that auto-regressive SNI lacks generalizability for diverse ASR systems.
If different ASR systems have similar error distributions, existing auto-regressive generation SNIs would generalize in an ASR zero-shot setting.
However, NoisyGen is consistently outperformed by NoisyGen (In Domain) in every task.
This result validates the ASR zero-shot setting where existing auto-regressive generation-based SNI models struggle to generalize in the other ASR systems.

Results of ISNI suggest that ISNI can robustify SLU model in the ASR zero-shot setting.
ISNI surpassed NoisyGen in every task in every noise level even NoisyGen (In Domain) in high and medium noise levels for QNLI and in low noise levels for SST2.
Such results might be attributed to the diversified ASR errors in the ASR GLUE benchmark, which ISNI is specifically designed to target.

ISNI demonstrated robust performance across varying noise levels compared to baselines, maintaining its efficacy as noise levels escalated.
This result highlights that ISNI can better robustify against phoneme confusions, which increase under noisy conditions~\cite{wu2022phonetic,petkov2013maximizing}.




\subsection{RQ2: Robustification of ISNI on Various SLU Tasks.}
We demonstrate that ISNI significantly enhances robustness across various SLU tasks in an ASR zero-shot setting, particularly in KS, where lexical perturbation heavily influences retrieval~\cite{ir_error_robustness1, ir_error_robustness2}.

Results from the DSTC10 Track 2 dataset, in Table~\ref{tab:dstc_results}, 
reveal that while baseline models struggle in the ASR zero-shot setting, showing R@1 score below 60, ISNI consistently outperforms these models across all metrics.
This superior performance, especially notable in KS, validates ISNI's effectiveness against errors from unknown ASR systems, unlike previous models that require identical ASR systems for both training and testing~\cite{asrglue,noisygen}.

Additionally, the robustification of ISNI extends beyond KS.
ISNI excels across all evaluated tasks, markedly improving the BLEU score by 2-3 times for Response Generation (RG). 
These findings affirm ISNI's capacity to substantially mitigate the impact of ASR errors across diverse SLU tasks.

\subsection{RQ3: Importance of Each Proposed Method to the Robustification. }
\begin{table}[h]
\centering
\scalebox{0.7}{
\begin{tabular}{l|c|c|c|c}
\hline
Noise Level                & High  & Medium & Low   & Clean \\ \hline
NoisyGen                   & 46.43 & 50.00  & 58.33 & 60.71 \\ \hline
NoisyGen (In Domain)       & 53.57 & 58.93  & 59.52 & 60.71 \\ \hline
ISNI (Ours)                & 56.55 & 58.93  & 60.12 & 60.71 \\
- phoneme aware generation & 53.57 & 55.35  & 60.12 & 61.31 \\
\quad - intervention             & 52.38 & 54.17  & 60.12 & 60.43 \\ \hline
\quad \quad -phoneme similarity loss   & 51.19    & 51.78     & 58.93    & 60.72    \\ \hline
\end{tabular}
}
\label{tab:rteablate}
\caption{Ablation study on RTE of ASR GLUE benchmark.}
\end{table}
To evaluate the contribution of each component in ISNI, we performed ablation studies on both the DSTC10 Track2 dataset in Table~\ref{tab:dstc_results} and the RTE task of the ASR GLUE benchmark in Table 4. 

The results without the phoneme-aware generation are presented in the fourth row of Table~\ref{tab:dstc_results} and Table 4. 
Removing the phoneme-aware generation led to a drop in performance across all tasks in DSTC10 Track2, as well as at high and medium noise levels in the RTE task of the ASR GLUE benchmark. 
This result demonstrates how phoneme-aware generation improves the robustness of SLU models in noisy environments. By accounting for word pronunciation, it ensures pseudo transcripts are ASR-plausible.

We also conducted an ablation study on the intervention by removing the do-calculus-based random corruption. 
For such goal, we trained a constrained decoding-based SNI model without \textit{do}-calculus-based random corruption as in Eq.~\ref{eq:do-calculus}.
For the constrained decoding method in ASR correction~\cite{constdecoder}, the corruption module, which determines whether the word $x^k$ will be corrupted, is jointly learned with the generation decoder.
Such module can be considered as implementing $P(z|x^k)$ in Eq.~\ref{eq:likelihood}.

The results, shown in the fifth row of Table~\ref{tab:dstc_results} and Table 4, indicate a performance decrease, particularly in the KS subtask of DSTC10 Track2 and in the RTE task, which are largely influenced by lexical perturbations due to  ASR errors. 
A similar decline is observed in the RG subtask, further emphasizing the importance of the intervention in generating diverse and generalized ASR errors. 
However, we noticed a slight performance increase in the KTD task. 
We hypothesize that this improvement may be attributed to the nature of text classification tasks, where robustness against minor lexical changes may not be as critical. 
Despite this, previous research suggests that abundant noise in the training set can degrade text classification performance over time~\cite{4470224}, which ISNI is designed to mitigate.

Finally, we performed an ablation study on the phonetic similarity loss by training ISNI without phonetic similarity loss, relying solely on phoneme embeddings. 
The results, presented in the last row of Table 4, show a further reduction in performance across most noise levels.
By supervising the phonetic information, phonetic similarity loss ensures that the generated noise words remain phonetically realistic, which is essential for improving model robustness in noisy conditions.

\section{Conclusion}
In this paper, we address the challenge of ASR generalizability within the context of SNI. 
We focus on enhancing the robustness of SLU models against ASR errors from diverse ASR systems. 
Our contributions are two-fold: Firstly, ISNI significantly broadens the spectrum of plausible ASR errors, thereby reducing biases. 
Second, not to lose the generality to any ASR, we generate noises that are universally plausible, or ASR$_*$-plausible,
which is empirically validated through extensive experiments across multiple ASR systems.

\paragraph{Limitations.}
One limitation of ISNI is reducing the chance of making substitution errors for entire sequences of 2-3 words.
As in previous works in SNI, ISNI consumes one token at a time when producing outputs.
While such consumption does not restrict to make substitution errors for entire sequences, it may reduce the chance of making substitution errors for such long sequences, as ISNI decides to corrupt one token at a time. 
Secondly, as ISNI is trained with speech corpora collected for academic purposes, they may face challenges when adopted for real-world applications, including
diverse spoken language variations such as dialects and accents. 
These variations can introduce noises that are not phonetically similar, which are different from the speech data used during ISNI training. 
This discrepancy may cause ISNI to fail in robustifying SLU models as ISNI is not prepared to handle ASR errors from those speech variations.  
Addressing this limitation may require enlarging the training dataset for ISNI
to cover the diverse noises
from spoken language variations.



\section*{Acknowledgement}

This work was supported by Institute of Information \& communications Technology Planning \& Evaluation (IITP) grant funded by the Korea government(MSIT)  [NO.RS-2021-II211343, Artificial Intelligence Graduate School Program (Seoul National University)] and  Institute of Information \& Communications Technology Planning \& Evaluation (IITP) grant funded by the Korean government (MSIT)(No. 2022-0-00077/RS-2022-II220077, AI Technology Development for Commonsense Extraction, Reasoning, and Inference from Heterogeneous Data).
\bibliography{anthology,custom}

\begin{thebibliography}{36}
\expandafter\ifx\csname natexlab\endcsname\relax\def\natexlab#1{#1}\fi

\bibitem[{Agarwal et~al.(2007)Agarwal, Godbole, Punjani, and Roy}]{4470224}
Sumeet Agarwal, Shantanu Godbole, Diwakar Punjani, and Shourya Roy. 2007.
\newblock \href {https://doi.org/10.1109/ICDM.2007.21} {How much noise is too much: A study in automatic text classification}.
\newblock In \emph{Seventh IEEE International Conference on Data Mining (ICDM 2007)}, pages 3--12.

\bibitem[{Ahmed et~al.(2022)Ahmed, Suffian, Khan, and Bogliolo}]{9662078}
Tafseer Ahmed, Muhammad Suffian, Muhammad~Yaseen Khan, and Alessandro Bogliolo. 2022.
\newblock \href {https://doi.org/10.1109/ACCESS.2021.3137905} {Discovering lexical similarity using articulatory feature-based phonetic edit distance}.
\newblock \emph{IEEE Access}, 10:1533--1544.

\bibitem[{Belinkov and Bisk(2018)}]{belinkov2018synthetic}
Yonatan Belinkov and Yonatan Bisk. 2018.
\newblock \href {https://openreview.net/forum?id=BJ8vJebC-} {Synthetic and natural noise both break neural machine translation}.
\newblock In \emph{International Conference on Learning Representations}.

\bibitem[{Broscheit et~al.(2022)Broscheit, Do, and Gaspers}]{broscheit-etal-2022-distributionally}
Samuel Broscheit, Quynh Do, and Judith Gaspers. 2022.
\newblock \href {https://doi.org/10.18653/v1/2022.acl-long.139} {Distributionally robust finetuning {BERT} for covariate drift in spoken language understanding}.
\newblock In \emph{Proceedings of the 60th Annual Meeting of the Association for Computational Linguistics (Volume 1: Long Papers)}, pages 1970--1985, Dublin, Ireland. Association for Computational Linguistics.

\bibitem[{Chen et~al.(2024)Chen, Hu, Yang, Siniscalchi, Chen, and Chng}]{chen2024hyporadise}
Chen Chen, Yuchen Hu, Chao-Han~Huck Yang, Sabato~Marco Siniscalchi, Pin-Yu Chen, and Eng-Siong Chng. 2024.
\newblock Hyporadise: An open baseline for generative speech recognition with large language models.
\newblock \emph{Advances in Neural Information Processing Systems}, 36.

\bibitem[{Chen et~al.(2017)Chen, Hsu, Huang, and Lee}]{tts_asr2}
Pin-Jung Chen, I-Hung Hsu, Yi-Yao Huang, and Hung-Yi Lee. 2017.
\newblock Mitigating the impact of speech recognition errors on chatbot using sequence-to-sequence model.
\newblock In \emph{2017 IEEE Automatic Speech Recognition and Understanding Workshop (ASRU)}, pages 497--503. IEEE.

\bibitem[{Chen et~al.(2022)Chen, Luo, He, Sun, and Sun}]{ir_error_robustness2}
Xuanang Chen, Jian Luo, Ben He, Le~Sun, and Yingfei Sun. 2022.
\newblock \href {https://doi.org/10.24963/ijcai.2022/275} {Towards robust dense retrieval via local ranking alignment}.
\newblock In \emph{Proceedings of the Thirty-First International Joint Conference on Artificial Intelligence, {IJCAI-22}}, pages 1980--1986. International Joint Conferences on Artificial Intelligence Organization.
\newblock Main Track.

\bibitem[{Cui et~al.(2021)Cui, Xiao, Li, Jiang, and Liu}]{noisygen}
Tong Cui, Jinghui Xiao, Liangyou Li, Xin Jiang, and Qun Liu. 2021.
\newblock \href {http://arxiv.org/abs/2103.13610} {An approach to improve robustness of nlp systems against asr errors}.

\bibitem[{Devlin et~al.(2019)Devlin, Chang, Lee, and Toutanova}]{devlin-etal-2019-bert}
Jacob Devlin, Ming-Wei Chang, Kenton Lee, and Kristina Toutanova. 2019.
\newblock \href {https://doi.org/10.18653/v1/N19-1423} {{BERT}: Pre-training of deep bidirectional transformers for language understanding}.
\newblock In \emph{Proceedings of the 2019 Conference of the North {A}merican Chapter of the Association for Computational Linguistics: Human Language Technologies, Volume 1 (Long and Short Papers)}, pages 4171--4186, Minneapolis, Minnesota. Association for Computational Linguistics.

\bibitem[{Di~Gangi et~al.(2019)Di~Gangi, Enyedi, Brusadin, and Federico}]{di-gangi-etal-2019-robust}
Matti Di~Gangi, Robert Enyedi, Alessandra Brusadin, and Marcello Federico. 2019.
\newblock \href {https://aclanthology.org/2019.iwslt-1.32} {Robust neural machine translation for clean and noisy speech transcripts}.
\newblock In \emph{Proceedings of the 16th International Conference on Spoken Language Translation}, Hong Kong. Association for Computational Linguistics.

\bibitem[{Dutta et~al.(2022)Dutta, Jain, Maheshwari, Ramakrishnan, and Jyothi}]{robart}
Samrat Dutta, Shreyansh Jain, Ayush Maheshwari, Ganesh Ramakrishnan, and Preethi Jyothi. 2022.
\newblock \href {http://arxiv.org/abs/2202.01157} {Error correction in {ASR} using sequence-to-sequence models}.
\newblock \emph{CoRR}, abs/2202.01157.

\bibitem[{Feng et~al.(2022)Feng, Yu, Wang, Liu, Cai, and Zheng}]{asrglue}
Lingyun Feng, Jianwei Yu, Yan Wang, Songxiang Liu, Deng Cai, and Haitao Zheng. 2022.
\newblock \href {https://doi.org/10.21437/Interspeech.2022-10097} {{ASR-Robust Natural Language Understanding on ASR-GLUE dataset}}.
\newblock In \emph{Proc. Interspeech 2022}, pages 1101--1105.

\bibitem[{Gopalakrishnan et~al.(2020)Gopalakrishnan, Hedayatnia, Wang, Liu, and Hakkani-Tür}]{confusion_matrix2}
Karthik Gopalakrishnan, Behnam Hedayatnia, Longshaokan~Marshall Wang, Yang Liu, and Dilek Hakkani-Tür. 2020.
\newblock \href {https://www.amazon.science/publications/are-neural-open-domain-dialog-systems-robust-to-speech-recognition-errors-in-the-dialog-history-an-empirical-study} {Are neural open-domain dialog systems robust to speech recognition errors in the dialog history? an empirical study}.
\newblock In \emph{Interspeech 2020}.

\bibitem[{Hannun et~al.(2014)Hannun, Case, Casper, Catanzaro, Diamos, Elsen, Prenger, Satheesh, Sengupta, Coates et~al.}]{deepspeech}
Awni Hannun, Carl Case, Jared Casper, Bryan Catanzaro, Greg Diamos, Erich Elsen, Ryan Prenger, Sanjeev Satheesh, Shubho Sengupta, Adam Coates, et~al. 2014.
\newblock Deep speech: Scaling up end-to-end speech recognition.
\newblock \emph{arXiv preprint arXiv:1412.5567}.

\bibitem[{Heigold et~al.(2018)Heigold, Varanasi, Neumann, and van Genabith}]{heigold-etal-2018-robust}
Georg Heigold, Stalin Varanasi, G{\"u}nter Neumann, and Josef van Genabith. 2018.
\newblock \href {https://aclanthology.org/W18-1807} {How robust are character-based word embeddings in tagging and {MT} against wrod scramlbing or randdm nouse?}
\newblock In \emph{Proceedings of the 13th Conference of the Association for Machine Translation in the {A}mericas (Volume 1: Research Track)}, pages 68--80, Boston, MA. Association for Machine Translation in the Americas.

\bibitem[{Huang and Chen(2020)}]{9054689}
Chao-Wei Huang and Yun-Nung Chen. 2020.
\newblock \href {https://doi.org/10.1109/ICASSP40776.2020.9054689} {Learning asr-robust contextualized embeddings for spoken language understanding}.
\newblock In \emph{ICASSP 2020 - 2020 IEEE International Conference on Acoustics, Speech and Signal Processing (ICASSP)}, pages 8009--8013.

\bibitem[{Jurafsky and Martin(2019)}]{jurafsky2019speech}
Dan Jurafsky and James~H Martin. 2019.
\newblock Speech and language processing (3rd (draft) ed.).

\bibitem[{Jyothi and Fosler-Lussier(2010)}]{confusion_matrix3}
Preethi Jyothi and Eric Fosler-Lussier. 2010.
\newblock Discriminative language modeling using simulated asr errors.
\newblock In \emph{Eleventh Annual Conference of the International Speech Communication Association}.

\bibitem[{Kim et~al.(2021)Kim, Liu, Jin, Papangelis, Gopalakrishnan, Hedayatnia, and Hakkani-Tur}]{dstc10_2}
Seokhwan Kim, Yang Liu, Di~Jin, Alexandros Papangelis, Karthik Gopalakrishnan, Behnam Hedayatnia, and Dilek Hakkani-Tur. 2021.
\newblock \href {http://arxiv.org/abs/2109.13489} {"how robust r u?": Evaluating task-oriented dialogue systems on spoken conversations}.

\bibitem[{Leng et~al.(2021)Leng, Tan, Wang, Zhu, Xu, Liu, Liu, Li, Qin, Lin, and Liu}]{leng-etal-2021-fastcorrect-2}
Yichong Leng, Xu~Tan, Rui Wang, Linchen Zhu, Jin Xu, Wenjie Liu, Linquan Liu, Xiang-Yang Li, Tao Qin, Edward Lin, and Tie-Yan Liu. 2021.
\newblock \href {https://doi.org/10.18653/v1/2021.findings-emnlp.367} {{F}ast{C}orrect 2: Fast error correction on multiple candidates for automatic speech recognition}.
\newblock In \emph{Findings of the Association for Computational Linguistics: EMNLP 2021}, pages 4328--4337, Punta Cana, Dominican Republic. Association for Computational Linguistics.

\bibitem[{Li and Specia(2019)}]{li-specia-2019-improving}
Zhenhao Li and Lucia Specia. 2019.
\newblock \href {https://doi.org/10.18653/v1/D19-5543} {Improving neural machine translation robustness via data augmentation: Beyond back-translation}.
\newblock In \emph{Proceedings of the 5th Workshop on Noisy User-generated Text (W-NUT 2019)}, pages 328--336, Hong Kong, China. Association for Computational Linguistics.

\bibitem[{Liu et~al.(2021)Liu, Takanobu, Wen, Wan, Li, Nie, Li, Peng, and Huang}]{tts_asr}
Jiexi Liu, Ryuichi Takanobu, Jiaxin Wen, Dazhen Wan, Hongguang Li, Weiran Nie, Cheng Li, Wei Peng, and Minlie Huang. 2021.
\newblock \href {https://doi.org/10.18653/v1/2021.acl-long.192} {Robustness testing of language understanding in task-oriented dialog}.
\newblock In \emph{Proceedings of the 59th Annual Meeting of the Association for Computational Linguistics and the 11th International Joint Conference on Natural Language Processing (Volume 1: Long Papers)}, pages 2467--2480, Online. Association for Computational Linguistics.

\bibitem[{Penha et~al.(2022)Penha, C\^{a}mara, and Hauff}]{ir_error_robustness1}
Gustavo Penha, Arthur C\^{a}mara, and Claudia Hauff. 2022.
\newblock \href {https://doi.org/10.1007/978-3-030-99736-6_27} {Evaluating the robustness of retrieval pipelines with query variation generators}.
\newblock In \emph{Advances in Information Retrieval: 44th European Conference on IR Research, ECIR 2022, Stavanger, Norway, April 10–14, 2022, Proceedings, Part I}, page 397–412, Berlin, Heidelberg. Springer-Verlag.

\bibitem[{Petkov et~al.(2013)Petkov, Henter, and Kleijn}]{petkov2013maximizing}
Petko~N Petkov, Gustav~Eje Henter, and W~Bastiaan Kleijn. 2013.
\newblock Maximizing phoneme recognition accuracy for enhanced speech intelligibility in noise.
\newblock \emph{IEEE transactions on audio, speech, and language processing}, 21(5):1035--1045.

\bibitem[{Ruan et~al.(2020)Ruan, Nechaev, Chen, Su, and Kiss}]{ruan20_interspeech}
Weitong Ruan, Yaroslav Nechaev, Luoxin Chen, Chengwei Su, and Imre Kiss. 2020.
\newblock \href {https://doi.org/10.21437/Interspeech.2020-2844} {{Towards an ASR Error Robust Spoken Language Understanding System}}.
\newblock In \emph{Proc. Interspeech 2020}, pages 901--905.

\bibitem[{Serai et~al.(2022)Serai, Sunder, and Fosler-Lussier}]{serai2022hallucination}
Prashant Serai, Vishal Sunder, and Eric Fosler-Lussier. 2022.
\newblock Hallucination of speech recognition errors with sequence to sequence learning.
\newblock \emph{IEEE/ACM Transactions on Audio, Speech, and Language Processing}, 30:890--900.

\bibitem[{Tam et~al.(2022)Tam, Xu, Zou, Wang, Liao, and Yuan}]{asr_pattern1}
Yik-Cheung Tam, Jiacheng Xu, Jiakai Zou, Zecheng Wang, Tinglong Liao, and Shuhan Yuan. 2022.
\newblock \href {https://doi.org/10.1109/icassp43922.2022.9746741} {Robust unstructured knowledge access in conversational dialogue with asr errors}.
\newblock In \emph{ICASSP 2022 - 2022 IEEE International Conference on Acoustics, Speech and Signal Processing (ICASSP)}. IEEE.

\bibitem[{Tian et~al.(2021)Tian, Huang, He, Lin, Bao, He, Huang, Ju, Zhang, Xie, Sun, Wang, Wu, and Wang}]{tian2021tod-da}
Xin Tian, Xinxian Huang, Dongfeng He, Yingzhan Lin, Siqi Bao, Huang He, Liankai Huang, Qiang Ju, Xiyuan Zhang, Jian Xie, Shuqi Sun, Fan Wang, Hua Wu, and Haifeng Wang. 2021.
\newblock Tod-da: Towards boosting the robustness of task-oriented dialogue modeling on spoken conversations.
\newblock \emph{arXiv preprint arXiv:2112.12441}.

\bibitem[{Tsvetkov et~al.(2014)Tsvetkov, Metze, and Dyer}]{tsvetkov-etal-2014-augmenting}
Yulia Tsvetkov, Florian Metze, and Chris Dyer. 2014.
\newblock \href {https://doi.org/10.3115/v1/E14-1065} {Augmenting translation models with simulated acoustic confusions for improved spoken language translation}.
\newblock In \emph{Proceedings of the 14th Conference of the {E}uropean Chapter of the Association for Computational Linguistics}, pages 616--625, Gothenburg, Sweden. Association for Computational Linguistics.

\bibitem[{Vaswani et~al.(2017)Vaswani, Shazeer, Parmar, Uszkoreit, Jones, Gomez, Kaiser, and Polosukhin}]{NIPS2017_3f5ee243}
Ashish Vaswani, Noam Shazeer, Niki Parmar, Jakob Uszkoreit, Llion Jones, Aidan~N Gomez, \L~ukasz Kaiser, and Illia Polosukhin. 2017.
\newblock \href {https://proceedings.neurips.cc/paper_files/paper/2017/file/3f5ee243547dee91fbd053c1c4a845aa-Paper.pdf} {Attention is all you need}.
\newblock In \emph{Advances in Neural Information Processing Systems}, volume~30. Curran Associates, Inc.

\bibitem[{Wang et~al.(2020)Wang, Fazel-Zarandi, Tiwari, Matsoukas, and Polymenakos}]{confusion_matrix}
Longshaokan~Marshall Wang, Maryam Fazel-Zarandi, Aditya Tiwari, Spyros Matsoukas, and Lazaros Polymenakos. 2020.
\newblock \href {https://www.amazon.science/publications/data-augmentation-for-training-dialog-models-robust-to-speech-recognition-errors} {Data augmentation for training dialog models robust to speech recognition errors}.
\newblock In \emph{ACL 2020 Workshop on NLP for Conversational AI}.

\bibitem[{Wu et~al.(2022)Wu, Lian, Jiang, Song, Zhao, Xu, and Yang}]{wu2022phonetic}
Xueyang Wu, Rongzhong Lian, Di~Jiang, Yuanfeng Song, Weiwei Zhao, Qian Xu, and Qiang Yang. 2022.
\newblock A phonetic-semantic pre-training model for robust speech recognition.
\newblock \emph{CAAI Artificial Intelligence Research}, 1(1):1--7.

\bibitem[{Yang et~al.(2022)Yang, Li, and Peng}]{constdecoder}
Jingyuan Yang, Rongjun Li, and Wei Peng. 2022.
\newblock \href {https://doi.org/10.21437/Interspeech.2022-660} {{ASR Error Correction with Constrained Decoding on Operation Prediction}}.
\newblock In \emph{Proc. Interspeech 2022}, pages 3874--3878.

\bibitem[{Yu et~al.(2016)Yu, Lee, and Lee}]{confusion_matrix5}
Lang-Chi Yu, Hung-yi Lee, and Lin-Shan Lee. 2016.
\newblock Abstractive headline generation for spoken content by attentive recurrent neural networks with asr error modeling.
\newblock In \emph{2016 IEEE Spoken Language Technology Workshop (SLT)}, pages 151--157. IEEE.

\bibitem[{Yuan et~al.(2017)Yuan, Xun, Suo, Jia, and Zhang}]{yuan2017wave2vec}
Ye~Yuan, Guangxu Xun, Qiuling Suo, Kebin Jia, and Aidong Zhang. 2017.
\newblock Wave2vec: Learning deep representations for biosignals.
\newblock In \emph{2017 IEEE International Conference on Data Mining (ICDM)}, pages 1159--1164. IEEE.

\bibitem[{Zhang et~al.(2019)Zhang, Zhang, Chen, and Zhang}]{8907842}
Zhichang Zhang, Zhenwen Zhang, Haoyuan Chen, and Zhiman Zhang. 2019.
\newblock \href {https://doi.org/10.1109/ACCESS.2019.2954766} {A joint learning framework with bert for spoken language understanding}.
\newblock \emph{IEEE Access}, 7:168849--168858.

\end{thebibliography}
\bibstyle{acl_natbib}
\section{Appendix}
\subsection{Related Works}
\label{sec:rel_work}
Previously proposed methods can be categorized into three categories. 
First, TTS (Text-to-Speech)-ASR pipeline~\cite{tts_asr, tts_asr2} adopts the TTS engine, the reverse of ASR, to convert written text into audio.
The ASR$_i$ transcribes it into pseudo transcription. 
However, the human recording and TTS-generated audio differ in their error distributions, which makes the resulting pseudo transcriptions different from ASR transcriptions~\cite{asrglue}. 

Second is textual perturbation, replacing the words $x^k \in X$ to the noise word $t_{i}^{k}$ as follows:
\begin{equation}
   t^{k}_i = argmax_{w \in W} \textit{S}_i(w|x^{k}),
\end{equation}
where $\textit{S}_i(w|x^{k})$ evaluates the score that ASR$_i$ would corrupt the written word $x^k$ into each word $w \in W$ in the vocabulary set $W$.
A widely adopted type of $\textit{S}_i(w|x^{k})$ is a confusion matrix built from the paired written and ASR transcribed corpora~\cite{confusion_matrix3, confusion_matrix5} or phonetic similarity function~\cite{li-specia-2019-improving,tsvetkov-etal-2014-augmenting}.

A representative of the above two categories is TOD-DA which embodies both categories. 

Third is  the auto-regressive generation, where
auto-regressive PLMs such as GPT-2 or BART are supervised to maximize the likelihood of the ASR transcription $T_i$ given its written text $X$. 
Adopting PLMs, the auto-regressive generation can consider contextual information and generate more ASR$_i$-plausible noise words~\cite{noisygen}.

Auto-regressive generation is biased to ASR$_i$, limiting the SLU model used for ASR$_i$.

Our distinction is generalizing SNI so that the SLU tasks can be conducted with ASR$_*$.

\subsection{Proof of Eq.~\ref{eq:do}}
\label{sec:proof_do}
Using the $do$ operator to the conventional likelihood, $P(Y|do(X))$ is transformed as follows:
\begin{align}
P(Y|do(X)) &= \sum_{z}P(Y, z|do(X))\\
&= \sum_{z}P(Y|do(X), z) \cdot P(z|do(X))\\
&= \sum_{z}P(Y|X, z) \cdot P(z|do(X))
\label{do_proof1}
\end{align}
Then, further transition is conducted by applying Rule 3 of $do$-calculus.
Rule 3 states that we can remove the $do$-operator if $X$ and $Z$ are independent in $G_{\bar{X}}$, a modified version of the causal graph of SNI where all arrows incoming to $X$ are removed.
In $G_{\bar{X}}$, where there are two paths, $X\rightarrow Y$ and $Z\rightarrow Y$, $X$, and $Z$ are independent, $X\perp Z$, as there is no valid path between $X$ and $Z$.
Removing the $do$-operator, Eq.~\ref{do_proof1} is evaluated as follows:
\begin{align}
P(Y|do(X))&= \sum_{z}P(Y|X, z) \cdot P(z).
\label{eq:do_proof2}
\end{align}

\subsection{ASR systems for SNI and Downstream Tasks.}
\label{sec:asr_details}
For training the SNI model, we used the open-source commercial ASR system, Mozilla DeepSpeech.
Mozilla DeepSpeech, which we adopted for SNI model training, is an RNN-based end-to-end ASR system trained on a 1700-hour audio dataset.
DeepSpeech shows similar word error rates to famous commercial ASR systems such as Google Translate’s speech-to-text API and IBM Watson Speech-to-text, in various benchmark datasets such as Librispeech clean test and Commonvoice as the table below shows. This similarity in performance makes it a relevant and practical choice for our study, providing a realistic and challenging testbed for our methods.
\begin{table}[h]
\scalebox{0.7}{
\begin{tabular}{l|ccc}
\hline
ASR                    & DeepSpeech & Google Translate & IBM Watson \\ \hline
Librispeech Clean  & 0.07       & 0.11             & 0.11       \\
CommonVoice            & 0.32       & 0.32             & 0.39       \\ \hline
\end{tabular}
}
\caption{WER of commercial ASR systems.}
\end{table}

Specifically, for the DSTC Track2 dataset, an unknown ASR system is used to generate transcriptions~\cite{dstc10_2}.
For the ASR GLUE benchmark, the LF-MMI TDNN ASR system is adopted to generate transcriptions~\cite{asrglue}.
ASR GLUE benchmark adopts an LF-MMI time-delayed neural network-based ASR syetem trained on a 6000-hour dataset.

\subsection{Dataset Details}
\label{sec:dataset_details}
\paragraph{SNI Training} For training SNI, we used the same datasets with our main baseline, Noisy-Gen~\cite{noisygen}, which used popular speech corpora, Common Voice, tatoeba audio, and LJSpeech-1.1 and MSLT, to collect ASR error transcriptions. 
Specifically, the audio recordings of the above corpora are fed to the DeepSpeech and get ASR transcriptions. 
Then, we compare ASR transcription with ground-truth transcription, ignoring punctuation and casing errors, so that only erroneous transcriptions would remain. 
Finally, we obtain about 930k pairs of ground-truth transcription and the ASR-noised transcription pair.
Among the resulting pairs, those from Common Voice, Tatoeba audio, and LJSpeech-1.1 are used for the training set and the others from MSLT are used for the validation set.

\paragraph{SLU Testing}
To show the ASR generalizability of the SNI models, we adopt two distinct SLU benchmarks, ASR GLUE and DSTC10 Track 2.
DSTC10 Track2 dataset models task-oriented dialogue systems with unstructured knowledge access in a spoken language.
We adopt the DSTC10 Track2 dataset as it covers various tasks in NLP in the dialogue domain where SLU is required frequently.
Specifically, it consists of three successive subtasks covering various SLU tasks: Knowledge-seeking Turn Detection (KTD), Knowledge Selection (KS), and Knowledge-grounded Response Generation (RG).
First, KTD aims to determine whether the dialogue turn requires external knowledge access or not as a classification. 
Once determined to require external knowledge, the second step is KS, which aims to retrieve the appropriate knowledge snippet by estimating the relevance between the given dialogue context and each knowledge snippet in the knowledge base.
Finally, the response is generated in RG, based on the dialogue context and the selected knowledge snippet. 

In the DSTC10 Track2 dataset, human responses are transcribed by an unknown ASR system.

Another benchmark, ASR GLUE is an SLU version of the widely adopted NLU benchmark, GLUE.
It provides the written GTs for the training set and the transcriptions of 3 noise levels spoken by 5 human speakers for the development set and test set.
As the ground-truth label for the test set is unavailable, we report the results on the development set and sample the validation set from the pseudo transcripts generated from the training set.
Among various subtasks, we provide the results of two NLI tasks, QNLI and RTE, and 1 sentiment classification task, SST, which the DSTC10 Track2 dataset does not contain.

\subsection{Details of ISNI model}
\label{sec:ISNI_model_detail}
We used Transformer decoder~\cite{NIPS2017_3f5ee243} with 1 layer which has 12 attention heads and 768 hidden layer dimensions.
We trained ISNI for 20 epochs with Adam optimizer with a learning rate of 0.00005.
Also, we set $\lambda_{w}$ and $\lambda_{phoneme}$ as 0.5 to balance the semantic and phonetic information.

\subsection{Quantitative study on the generated pseudo-transcripts. }
\label{sec:quantitiate_study}
In this section, we quantitatively study the characteristics of the pseudo-transcripts generated by our ISNI.
For this study, We generated pseudo transcripts in MSLT, our development set for SNI training.
We set $P(z)=0.45$ for pseudo transcripts generation, which is similar to the word error rate of ASR transcription as we will show in Table~\ref{tab:wer_results}.
We analyze the following characteristics:

\noindent \textbf{i)} How phonetically similar are our generated pseudo transcripts?

\noindent  \textbf{ii)} The word error rate.

\noindent \textbf{iii)} Which error types composes the noise in the generated pseudo transcripts?

First, we show that PLMs are insufficient to generate the ASR$_*$-plausible pseudo transcriptions.
The noise words would be ASR$_*$-plausible if it is phonetically similar to its written form as an ASR system would incorrectly transcribe into phonetically similar noise words.
Therefore, we compared the phoneme edit distance $\textit{D}(w^{p}, w^{q})$ between the written GT and the pseudo transcriptions.
For a fair comparison, we set all tokens to have identical $z$ for the noise word generation to ensure that the identical words are corrupted.

\begin{table}[h]
\centering
 \scalebox{0.75}{
\begin{tabular}{l|c}
\hline
Model      & PD ($\downarrow$)    \\ \hline
Ours                       & 62.02 \\
- phoneme-aware generation & 72.52 \\ \hline
\end{tabular}
}
\caption{Phoneme edit distance between generated pseudo transcriptions and written GTs (Lower is better).}
\label{tab4:phoneme_edit_distance}
\end{table}

Table~\ref{tab4:phoneme_edit_distance} shows that the pseudo transcriptions without phoneme-aware generation show 17\% larger phonetic distance.
This result shows that PLMs are not enough to generate the ASR$_*$-plausible pseudo transcriptions and ignorance of phonetic information is the obstacle to generating ASR$_*$-plausible pseudo transcriptions.

Then, we study the word error rate of the generated pseudo transcripts.
\begin{table}[h]
\centering
\begin{tabular}{l|c}
\hline
Model      & WER  \\ \hline
DeepSpeech & 0.46 \\
Ours       & 0.66 \\
Noisy-Gen   & 0.76 \\ \hline
\end{tabular}
\caption{Word error rate of the DeepSpeech transcriptions and the pseudo transcriptions generated by ours and Noisy-Gen.}
\label{tab:wer_results}
\end{table}
The results in Table~\ref{tab:wer_results} show that pseudo transcripts generated by our ISNI contain more errors than P(z).
However, compared to the baselines, the word error rate generated by our methods is more controlled, as we control whether to corrupt the word by P(z).

Then, we break down word error rate results by error types of insertion/deletion/substitutions.
\begin{table}[h]
\begin{tabular}{l|ccc}
\hline
error type & insertion & deletion & substituion \\ \hline
DeepSpeech & 0.20      & 0.29     & 0.51        \\
Ours       & 0.25      & 0.16     & 0.59        \\ \hline
\end{tabular}
\caption{Error type of the DeepSpeech transcriptions and pseudo transcripts generated by our ISNI.}
\label{tab:error_type_results}
\end{table}
Table~\ref{tab:error_type_results} shows that three error types take up similarly in DeepSpeech transcription and pseudo transcriptions by our ISNI.
This result shows that our ISNI can well handle three error types.

\subsection{Generated Pseudo Transcripts Examples}
\label{sec:pseudo_transcripts_examples}

We provide the examples of the generated pseudo transcripts in Table~\ref{tab:pseudo_transcript_example1} and \ref{tab:pseudo_transcript_example2}.
\begin{table}[h]
\begin{tabular}{l|cccc}
input             & \multicolumn{4}{c}{as  bestial}             \\ \hline
tokenized input   & as   & best   & \multicolumn{2}{c}{\#\#ial} \\ \hline
z                 & 0    & 0      & \multicolumn{2}{c}{1}       \\ \hline
generated output  & -    & -      & at       & \#\#ial <eos>        \\ \hline
pseudo transcript & \multicolumn{4}{c}{as best atial}         
\end{tabular}
\caption{Examples of pseudo transcripts generated by our ISNI.}
\label{tab:pseudo_transcript_example1}
\end{table}
Among the tokenized input, $z$ of $\#\#ial$ was set to 1, thus the constrained decoder generates its noise word. 
The constrained decoder made 1 substitution error $(bestial \rightarrow best)$ and 1 insertion error $(atail)$.  

\begin{table}[h]
  \scalebox{0.6}{
\begin{tabular}{l|cccccc}
input             & \multicolumn{6}{c}{only labored the gags}                                                                                               \\ \hline
tokenized input   & only & larbor & \#\#ed                                    & the                                    & gag & \#\#s                        \\ \hline
z                 & 0    & 0      & 1                                         & 1                                      & 0   & 1                            \\ \hline
generated output  & -    & -      & \#\#ed labor <eos> & the \#\#s <eos> & -   & <eos> \\ \hline
pseudo transcript & \multicolumn{6}{c}{only labored labor thes gag}                                                                                        
\end{tabular}
}
\caption{More examples of pseudo transcripts generated by our ISNI.}
\label{tab:pseudo_transcript_example2}
\end{table}

\subsection{Use of AI Assistants}
We used ChatGPT for grammatical corrections.

\end{document}